\documentclass[a4paper, fleqn]{cas-dc}

\usepackage{makecell}
\usepackage[authoryear]{natbib}

\usepackage{amsmath}
\usepackage{multicol}
\usepackage{multirow}
\usepackage{xcolor}
\usepackage{soul}
\def\tsc#1{\csdef{#1}{\textsc{\lowercase{#1}}\xspace}}
\tsc{WGM}
\tsc{QE}
\begin{document}
\let\WriteBookmarks\relax
\def\floatpagepagefraction{1}
\def\textpagefraction{.001}

% Short title
\shorttitle{Review of State-of-The-Art Methods for Java Code Generation from Natural Language Text}    

% Short author
%\shortauthors{J. Lopez Espejel et al.}  

% Main title of the paper
\title [mode = title]{A Comprehensive Review of State-of-The-Art Methods for Java Code Generation from Natural Language Text}  

% Title footnote mark
% eg: \tnotemark[1]
\tnotemark[1] 

% Title footnote 1.
% eg: \tnotetext[1]{Title footnote text}
%\tnotetext[1]{<tnote text>} 

% First author
%
% Options: Use if required
% eg: \author[1,3]{Author Name}[type=editor,
%       style=chinese,
%       auid=000,
%       bioid=1,
%       prefix=Sir,
%       orcid=0000-0000-0000-0000,
%       facebook=<facebook id>,
%       twitter=<twitter id>,
%       linkedin=<linkedin id>,
%       gplus=<gplus id>]

\author[1]{{Jessica} {López Espejel}}[
    orcid=0000-0001-6285-0770
]

% Corresponding author indication
\cormark[1]

% Footnote of the first author
%\fnmark[1]

% Email id of the first author
\ead{jlopezespejel@novelis.io}

% URL of the first author
%\ead[url]{<URL>}

% Credit authorship
% eg: \credit{Conceptualization of this study, Methodology, Software}
%\credit{<Credit authorship details>}

% Address/affiliation
\affiliation[1]{organization={Novelis
Research and Innovation Lab},
            addressline={207 Rue de Bercy}, 
            city={Paris},
%          citysep={}, % Uncomment if no comma needed between city and postcode
            postcode={75012}, 
            state={},
            country={France}}

\author[1]{{Mahaman Sanoussi} {Yahaya Alassan}}[]

% Footnote of the second author
%\fnmark[1]

% Email id of the second author
\ead{syahaya@novelis.io}

% URL of the second author
%\ead[url]{}

% Credit authorship
\credit{}

% Address/affiliation
%\affiliation[1]{organization={},
%            addressline={}, 
%            city={},
%          citysep={}, % Uncomment if no comma needed between city and postcode
%            postcode={}, 
%            state={},
%            country={}}

\author[1]{El Mehdi Chouham}[]
% Footnote of the third author
%\fnmark[1]

% Email id of the second author
\ead{elchouham@novelis.io}

% URL of the second author
%\ead[url]{}

% Credit authorship
\credit{}

\author[1]{Walid Dahhane}[orcid=0000-0001-5387-3380]
% Footnote of the third author
%\fnmark[1]

% Email id of the second author
\ead{wdahhane@novelis.io}

% URL of the second author
%\ead[url]{}

% Credit authorship
\credit{}

% Address/affiliation
%\affiliation[1]{organization={},
%            addressline={}, 
%            city={},
%          citysep={}, % Uncomment if no comma needed between city and postcode
%            postcode={}, 
%            state={},
%            country={}}

\author[1]{El Hassane Ettifouri}[
    orcid=0000-0001-5299-9053
]
% Footnote of the fourth author
%\fnmark[1]

% Email id of the second author
\ead{eettifouri@novelis.io}

% URL of the second author
%\ead[url]{}

% Credit authorship
\credit{}

% Address/affiliation
%\affiliation[1]{organization={},
%            addressline={}, 
%            city={},
%          citysep={}, % Uncomment if no comma needed between city and postcode
%            postcode={}, 
%            state={},
%            country={}}

% Corresponding author text
\cortext[1]{Corresponding author}

% Footnote text
\fntext[1]{}
% Here goes the abstract
\begin{abstract}
    Java Code Generation consists in generating automatically Java code from a Natural Language Text. This NLP task helps in increasing programmers' productivity by providing them with immediate solutions to the simplest and most repetitive tasks. Code generation is a challenging task because of the hard syntactic rules and the necessity of a deep understanding of the semantic aspect of the programming language. Many works tried to tackle this task using either RNN-based, or Transformer-based models. The latter achieved remarkable advancement in the domain and they can be divided into three groups: (1) encoder-only models, (2) decoder-only models, and (3) encoder-decoder models.    
    In this paper, we provide a comprehensive review of the evolution and progress of deep learning models in Java code generation task. We focus on the most important methods and present their merits and limitations, as well as the objective functions used by the community. 
    In addition, we provide a detailed description of datasets and evaluation metrics used in the literature. Finally, we discuss results of different models  on CONCODE dataset, then propose some future directions.  % I do not have more ideas, I will think about later
\end{abstract}

\begin{graphicalabstract}
\includegraphics[width=\textwidth,trim={0cm 8cm 0cm 1cm}]{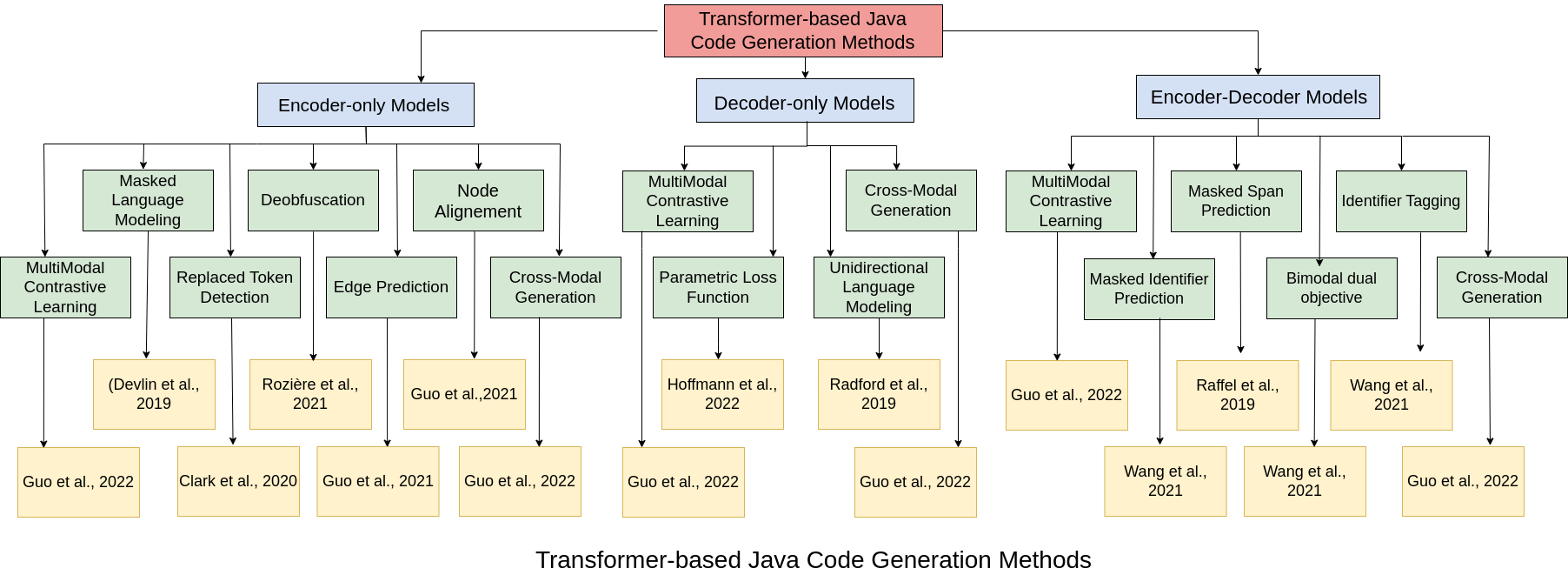}
\end{graphicalabstract}

% Research highlights
\begin{highlights}
\item Code Generation is an important Natural Language Processing (NLP) task
\item Initializing models from pretrained weights leads to better results than training them from scratch
\item  Decoder-only models achieve the best comprehension of the code syntax and semantics
%\item Masked Span Prediction (MSP) is the most crucial objective so far to learn the syntactic information in the generation tasks
\item Combining multiple learning objectives lead to better code generation models.
\item Improving Code Generation metrics is becoming a must in order to better compare and further improve state-of-the-art methods
\end{highlights}

% Keywords
% Each keyword is seperated by \sep

\ExplSyntaxOn
\keys_set:nn { stm / mktitle } { nologo }
\ExplSyntaxOff

\maketitle

% Main text
\section{Introduction}
\label{sec: introduction}
    
    In the last years, there has been a huge interest from the Natural Language Processing (NLP) community in the automation of software engineering to increase programmers' productivity \citep{ahmad-etal-2021-PLBART}. Code Generation Task (or \textit{Program Synthesis Task}) helps considerably in reducing workload of programmers, by  speeding up the implementation of simple functions, and letting them focus on the most complex tasks only. 
    
\begin{figure}[h!]
        \centering
        \includegraphics[width=0.47\textwidth]{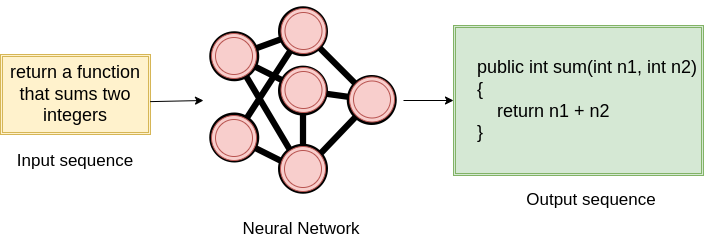}
        \caption{Java source code generation from natural language}
        \label{fig:text2java}
    \end{figure}

    In this paper, we are interested to show the progress over the years in the automatic Java source code generation from natural language.  This task consists in taking as an input a natural language phrase, and producing an equivalent Java code as an output (Figure \ref{fig:text2java}). 

    In the Java Code Generation task as in many others, attention is put on both the model and the dataset. The latter are key ingredients to develop a powerful Artificial Intelligence (AI) system. Intuitively, most of code generation datasets are built from source code of the concerned Programming Language (PL) and documentation in Natural Language (NL). These datasets are collected from web sites such as Github\footnote{\url{https://github.com/}} and Stack Overflow\footnote{\url{https://stackoverflow.com/}}. In the case of Java programming language,  documentation  is written in JavaDoc \citep{JavaDoc_web}, which is a format that describes all classes, member variables and methods. Therefore, JavaDoc documentation with its source code have been exploited to generate large datasets containing pairs of text-code lines (See Section \ref{sec:datasets} for more details). Along with datasets, several models have surged to tackle the Java Code Generation task. The first models used in literature are based on Recurrent Neural Networks (RNNs) such as LPN \citep{Ling016_LatentPredictor} and Seq2Seq \citep{yin2017_syntactic}. However, the big success of Transformer-based pretrained language models (LMs) such as BERT \citep{Devlin2019_BERT}, GPT \citep{radford2019_gpt2}, and T5 \citep{Colin_2020t5}, in a wide range of NLP tasks, shifted the efforts of the community to adapt Transformers models \citep{vaswani2017_attention} to generate Java code from natural language. The results of such efforts gave birth to more sophisticated Large Language Models (LLMs) such as ChatGPT \citep{chatGPT4}, LLaMA \citep{touvron2023_llama}, and BARD \citep{Google_Bard}. Note that at the time of writing this article, the performance of the latter algorithms has not been tested on standard Java benchmarks. Nonetheless, initial indications suggest that their performance is exceptional.

    Programming language generation presents more challenges than standard natural language generation. For instance, (1) the neural network should be able to correctly understand instructions in natural language to generate a corresponding correct source code, (2) the latter has lexical, grammatical, and semantic constraints \citep{wang-etal-2021-codet5, Scholak2021_PICARD} that should be taken into account, and (3) one small mistake (such as a missing dot or colon) can change completely the semantic of the code and make the model's output incorrect.

    In this paper, we provide a comprehensive review of state-of-the-art methods in Java code generation from natural language. To the best of our knowledge, this is the first review for Java Code Generation methods. Our goal is to provide a solid base for future researchers by highlighting what was already done, and what can be improved. These are the main findings of our paper:
    \begin{itemize}
        \item Initializing models from pretrained weights leads to better results than training them from scratch
        \item  Decoder-only models have shown the best comprehension of the code syntax and semantics
        %\item Masked Span Prediction (MSP) is the mostcrucial objective so far to learn the syntactic information in the generation tasks
        \item Combining multiple learning tasks lead to better code generation models.
        \item Improving Code Generation metrics is becoming a must in order to better compare and further improve state-of-the-art methods
    \end{itemize}

    The rest of the paper is organized as follows: Section \ref{sec:problem-definition} provides a formulation of the problem. Section \ref{sec:background} outlines the background of the studied task. Section \ref{sec:rnn_based} presents RNN-based methods to tackle Java Code Generation. Note that this type of methods are outdated nowadays and are not effective to tackle complex NLP tasks. Section \ref{sec:transformer_based} presents  the most powerful Transformer-based methods. This section is the largest, and takes the biggest attention in this article. Section \ref{sec:datasets} provides a brief description of the most popular datasets used in Java Generation. Section \ref{sec:performanceMetrics} presents evaluation metrics used by the community to evaluate different models. In Section \ref{sec:results}, we compare experimentally between the state-of-the-art methods and highlight their advantages and disadvantages. Finally, we conclude in Section \ref{sec:Conclusions} and provide some perspectives.

\section{Problem Definition}
\label{sec:problem-definition}

    We build on the notations used in \cite{feng2020_codebert} and propose a unified notation for all methods described in this paper. The goal is to facilitate the comprehension and the comparison between state-of-the-art methods.

    Given an input text $w = \{w_0, w_1,.., w_{n-1}\}$ of length $|w| = n$, the goal of a code generation model  is to produce an equivalent source code sequence $c = \{c_0, c_1,.., c_{l-1}\}$ of length $|c| = l$, where $w_i$ and $c_i$ are input and output (target) tokens respectively. The model is trained on batches of size $B$ using a loss $\mathcal{L}(\theta)$, where $\theta$ are the model's parameters. 
    The formula of loss function $\mathcal{L}(\theta)$ varies between code generation methods and depends heavily on the architecture used. We will provide a detailed explanation of these learning objectives in Section \ref{sec:transformer_based}.  
    
    \paragraph{\textbf{Data type -}} Code generation models  can be trained on unimodal or bimodal data. On the one hand,  unimodal data keep the source code without paired natural language, or keep the natural language without paired code. On the other hand,  bimodal data have NL-PL pairs. The bimodal data have shown its usefulness in training the multi-modal pretrained models \citep{feng2020_codebert}, which learn implicit alignment between input of different categories.

    \begin{figure*}[!t]
        \centering
        \includegraphics[width=0.95\textwidth]{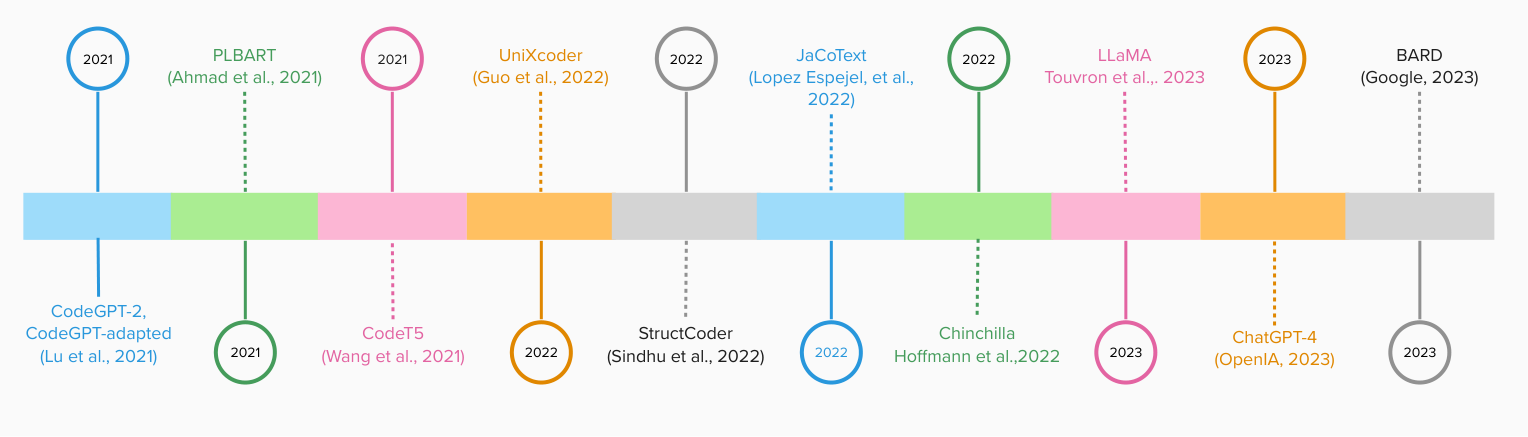}
        \caption{Timeline of some models in the Java Code Generation task}
        \label{fig:timeline}
    \end{figure*}

    \section{Background} 
    \label{sec:background}

    The automatic source code generation task can be modeled as a translation task, where the input text is a natural language phrase and the output text is a programming source code. Early works in the field of language translation focused on regular expressions, logical forms, sequence of instructions, and agent-specific languages. Later, the efforts were shifted to use semantic parsing and SQL code generation. These developments paved the way for the generation of Java code for natural language. In the following paragraphs, we will provide a brief overview of remarkable methods proposed for each category.

    \vspace{-1em}

     \paragraph*{\textbf{Regular expressions -}} \cite{Angluin1987_LearningRegular} worked on the mapping of natural language to regular expressions by learning general rules from examples, and automatizing their generation. Alternatively, \cite{Ranta1998_Multilingual} and \cite{kushman2013_UsingSemantic} used rule-based techniques and Combinatory Categorical Grammar (CCG) to generate regular expressions from natural language, respectively. Similarly, \cite{kushman2013_SemanticUnification} worked on generating regular expressions from natural language queries. While the method proved effective in handling a wide variety of natural language queries, including complex ones, it was not able to handle all of them, as it relies on semantic unification and could struggle with certain types of queries. 

     \vspace{-1em}

     \paragraph*{\textbf{Logical forms -}} Authors of \cite{Zettlemoyer2005_LearningToMap} mapped natural language sentences to logical forms. They introduced an approach based on probabilistic categorial grammars (PCGs). Even though the method handles ambiguity in natural language, and can generate multiple logical forms, the PCG parser used in the model is complex and computationally expensive, making it difficult to scale to large datasets.

     \vspace{-1em}

     \paragraph*{\textbf{Agent-specific language -}}  \cite{Kate2005_LearningTransform} proposed a method for automatically translating natural language sentences into agent-specific language expressions. The advantages of this method is that it can automate the translation process, saving time and effort for human translators, and can potentially improve the accuracy of formal language expression generation. However, the model could face difficulties in processing certain categories of complex sentences or in handling expressions that have multiple valid formal language translations.

    \vspace{-1em}

     \paragraph*{\textbf{Sequence of instructions -}} \cite{branavan2009_reinforcement} presented a framework that learns to map natural language instructions to corresponding actions in an environment by training an agent with a combination of supervised learning and reinforcement learning techniques. The method does not require hand-engineered features, making it highly adaptable to different tasks and environments. However, the framework struggles with instructions that are ambiguous or imprecise, leading to errors in action generation.

\vspace{-1em}

     \paragraph*{\textbf{Semantic Parsing -}}  We have identified salient background approaches in the semantic parsing. One example is the work introduced by \cite{Zettlemoyer2005_LearningToMap}, who mapped NL to lambda–calculus encodings  of their semantics. To check the syntax and the semantics, the authors used a log-linear model. Later, \cite{wong2006_LearningSemantic}, developed a statistical semantic parsing system called WASP. This system is designed to generate a formal representation of the meaning of a sentence. Similarly, \cite{lu2008_generativeModel} proposed for parsing natural language, which involves utilizing a generative model to convert input sentences into representations of their meanings.

\vspace{-1em}

     \paragraph*{\textbf{SQL query generation -}} This is the closest type of approaches to Java generation task. One interesting work in this category was proposed by  \cite{miller1996_FullyStatistical}  who used an interface that is based on trained statistical models. To train the model, authors used ATIS (Air Travel Information) dataset. This dataset also was used by \cite{Ramaswamy2000_HierarchicalFT}.

    The aforementioned works offer an introduction to the methods utilized in the background and establish the groundwork for more advanced techniques based on RNN and Transformers, which will be discussed in subsequent sections of this paper. Unfortunately, most of the methods presented in this section are outdated and are unable to deal with complex Java generation programs.

    \section{RNN-based Code Generation Methods} 
    \label{sec:rnn_based}

    The first neural-network-based approaches used to tackle source code generation from natural language were based on Recurrent Neural Networks (RNNs) such as Long Short-term Memory (LSTM)  \citep{hochreiter1997_lstm} and Gated Recurrent Unit (GRU) \citep{cho2014_gru}. For instance, \cite{Neelakantan2016_NeuralProgrammer} proposed an architecture known as the Neural Programmer, which integrates RNNs with a set of fundamental arithmetic and logic operations to enhance program induction methods. One key feature of their approach is the integration of additional memory into neural networks \citep{graves2014_neuralTuring,kumar2015_askAnything,Joulin2016_InferringAlgorithmic}, which allows for more advanced and sophisticated problem-solving capabilities. Moreover, \cite{Mou2015_End-to-End} used a method based on RNNs for generating computer programs directly from natural language input. This sequence-to-sequence model is an end-to-end approach that uses an LSTM-based encoder to create a fixed-length vector representation of the input, and a decoder that predicts each token based on the output  sequence tokens and the input vector.

    Similarly to \cite{Neelakantan2016_NeuralProgrammer}, \cite{Scott2016_NeuralProgrammerInterpreters} introduced an approach to induce programs called NPI (Neural Programmer-Interpreter). As \cite{Mou2015_End-to-End}, NPI is an end-to-end model. NPI does not need manual feature engineering or domain-specific knowledge. It can learn how to create and execute programs, but the method is limited to generate complex programs, as the size of the programs is restricted by the capacity of the neural network. In the same year, \cite{Ling016_LatentPredictor} proposed Latent Predictor Network (LPN) that generates code from natural language descriptions by predicting the latent representation of code snippets. The LPN model consists of two components: a code generation model and a latent predictor model. The code generation model maps natural language descriptions to code representations, and the latent predictor model predicts the latent code representation given the previous generated code. However, LPN model requires a large amount of training data to perform well.

    Another interesting work in this category is called DeepAPI. It was proposed by \cite{Gu2016_DL-API}, and consists in generating Java API sequences from natural language query. DeepAPI adapts a RNN encoder-decoder model \citep{cho2014_learning} to encode the input sequence to a fixed-length context vector. This helps in recognizing semantically related words. In addition, importance of individual APIs is used to enhance the performance of the model. DeepAPI is the first model from its kind, but unfortunately, it only focuses on JDK library.

    \cite{iyer-etal-2018-concode} introduced a sequence-to-sequence model and evaluated its performance on their CONCODE dataset. The model is composed of a bidirectional LSTM encoder, and an LSTM-based decoder. The latter uses a two-step attention mechanism, one of these steps is a supervised copy of \cite{Gu2016_CopyingMechanism}. Alternatively, \cite{iyer-etal-2018-concode} tested seq2seq model introduced by \cite{yin2017_syntactic} on their CONCODE dataset. Seq2seq uses sequence-to-sequence LSTM-based architecture. However, the decoder uses supervised copying from the whole input sequence.

    Despite the usefulness of RNNs in multiple NLP tasks, they are currently not very popular in the community since they were highly outperformed in 2017 by Transformer neural networks \citep{vaswani2017_attention}. Most of the works presented in this section did not focus on Java programming language. This is showed in Figure~\ref{fig:timeline} that outlines some methods focus on Java code generation. We can clearly see that Java PL task is indeed tackled using  Transformers. For this reason, we booked more room for them in the rest of the paper.

\section{Transformer-based Java Code Generation Methods} 
\label{sec:transformer_based}

\begin{figure}
        \centering
        \includegraphics[width=0.52\textwidth]{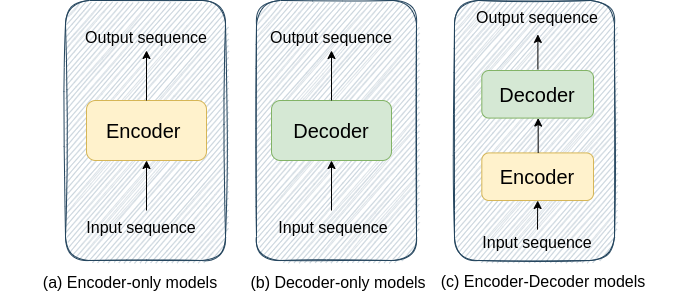}
        \caption{A simplified illustration of the different architecture types of Transformer neural networks}
        \label{fig:transformer_architecture}
    \end{figure}

\begin{figure*}[h!]
       \centering
       \includegraphics[width=\textwidth]{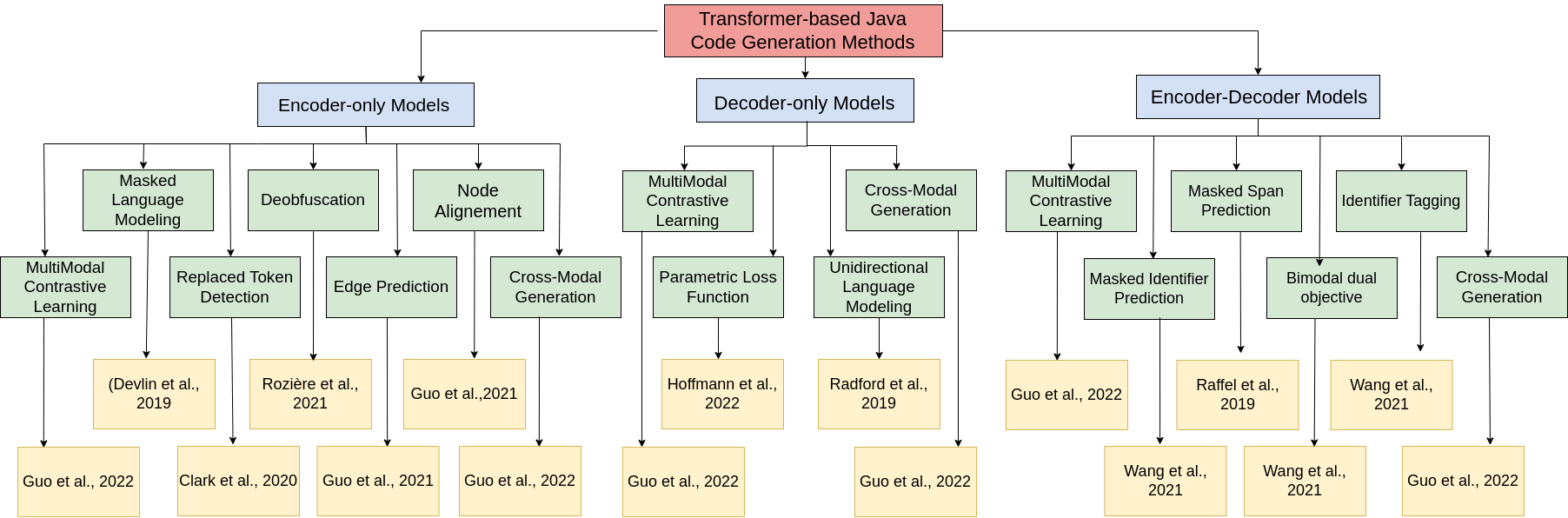}
       \caption{Overview of learning objectives in Transformer-based Code Generation Methods, and some examples of such methods.}
       \label{fig:transformer_category}
   \end{figure*}

    Transformers-based models have shown outstanding scores on various NLP tasks such as translation \citep{vaswani2017_attention}, automatic summarization \citep{Zhang2020_PEGASUS} and question answering \citep{Colin_2020t5}. All of these models have two stages: (1) pre-training, and (2) fine-tuning. The pre-training consists of learning a language model or a learning objective task from unlabeled data, and fine-tuning is the step in which the model learns the knowledge that is related to a specific task using labeled data. Therefore, one of the main factors in the success of these models is the supervised learning objectives used. Code generation methods tried first to adapt existing standard text generation methods and their objectives to Code Generation \citep{phan-etal-2021-cotext, ahmad-etal-2021-PLBART}. Later, other methods explored more specific techniques to improve the syntax and semantics of the generated code \citep{sindhu-etal-2022-StructCoder, wang-etal-2021-codet5}. We study state-of-the-art methods based on their architecture and divide then into three main categories:  encoder-only models, decoder-only models, and encoder-decoder models. Figure \ref{fig:transformer_architecture} shows a simplified illustration of the main difference in the architecture between the three types of models. Figure \ref{fig:transformer_category} shows the main learning objectives used in each category of methods, plus, an example of reference for each objective. We provide more details  in the following three subsections.

    \subsection{Encoder-only models}
    \label{subsec:encoder_only}

    These methods use an encoder only in the Transformer neural network architecture (Figure \ref{fig:transformer_architecture} (a)). They concatenate the natural language sequence $w$, and the programming language sequence $c$ and use them as an input $x = \{w, c\}$ to the encoder. 
    
    The main objectives used by this type of methods are as follows:

        \paragraph{$\bullet$ MLM: Masked Language Modeling} \citep{Devlin2019_BERT} - It consists in selecting randomly a set of positions from the input tokens, then replacing tokens in these positions with a special~\texttt{[MASK]} token. This objective aims to predict only the original tokens that are masked, and not to reconstruct the entire input. The token chosen at each position is selected from the model vocabulary. The MLM loss function is defined in Equation~\ref{eq:MLM}.

        \begin{equation}
            \mathcal{L}_{MLM}(\theta) = -\sum_{i \in m^{w} \cup m^{c}} log~P_\theta^{D_{1}} (x_{i} | w^{masked}, c^{masked})
            \label{eq:MLM}
        \end{equation}

            where $\theta$ are the model's parameters, $w$ is a natural language input, $c$ is programming language output, $P_\theta^{D_{1}}$ is the discriminator that predicts a token from a large vocabulary, and $m^w$ and $m^c$ are random set of masked positions in $w$ and $c$, respectively. 
        
       \paragraph{$\bullet$ RTD: Replaced Token Detection} \citep{Clark2020_ELECTRA} - It replaces (corrupts) some tokens in the input sequence with plausible tokens proposed from a generator network. The discriminator neural network is trained to determine if each token in the corrupt input is the original one or not. The RTD loss function is defined in Equation~\ref{eq:RTD}.

          \begin{equation}
              \begin{array}{l}
                \mathcal{L}_{RTD}(\theta) = \sum_{i=0}^{n+l-1} (\delta(i)~log~P_\theta^{D_{2}} (x^{corrupt}, i) + \\ 
                (1 - \delta(i)) (1 -log~P_\theta^{D_{2}} (x^{corrupt}, i))) 
              \end{array}
              \label{eq:RTD}
        \end{equation}

        where $P_\theta^{D_{2}}$ is the discriminator which predicts the probability of the $i$-th word being original, and  $\delta(i)$ is an indicator function defined as: 
            \begin{displaymath}
            \delta(i) = \left\{ \begin{array}{ll}
            1 & \mbox{$if$ $x_{i}^{corrupt} \equiv x_{i}$,} \\
            0 & \mbox{$otherwise$.}
            \end{array}
            \right.
            \end{displaymath}

        \paragraph{$\bullet$ DOBF: Deobfuscation}  \citep{Baptiste2021_DOBF} - It renames the class, function and variable names with uninformative labels. All instances of  selected identifiers are replaced in the whole source code with the uninformative label. The model is trained to  recover the original names by deobfuscating the code. It is the first approach to exploit the structure of programming languages.
    
        \paragraph{$\bullet$ Edge Prediction} \citep{Guo2021_GraphCodeBERT} - It is used with Data Flow Graph (DFG) and consists in masking some direct edges in the data flow. Then, the model aims to predict these masked edges. Edge Prediction loss is defined in Equation~\ref{eq:edge-prediction}.

        \begin{equation}
          \begin{array}{l}
            \mathcal{L}_{Edge Pred}(\theta) = - \sum_{e_{ij} \in E_{c}}^{} [ \delta(e_{ij} \in E^{masked})~log~p_{e_{ij}} + \\ 
            (1 - \delta(e_{ij} \in E^{masked}))~log~(1 - p_{e_{ij}})]
          \end{array}
          \label{eq:edge-prediction}
        \end{equation}

    where $E_c$ is the set of candidate edges,  $E^{masked}$ is the set of masked edges, $p_{e_{ij}}$ is the probability that an edge exists between nodes $i$ and $j$, and $\delta(e_{ij} \in E)$ is defined as follows:
            
        \begin{displaymath}
            \delta(e_{ij} \in E) = \left\{ \begin{array}{ll}
            1 & \mbox{$if~\langle v_i, v_j \rangle \in E$}, \\
            0 & \mbox{$otherwise$.}
            \end{array}
            \right.
        \end{displaymath}    
       
    \paragraph{$\bullet$ Node Alignement} \citep{Guo2021_GraphCodeBERT} -  It is similar to the Edge Prediction task, but instead of predicting edges between the nodes of DFG, it predicts edges between the source code tokens and  data flow nodes. 

     \paragraph{\textbf{\underline{Methods} -}}  CodeBERT \citep{feng2020_codebert} is based on BERT \citep{Devlin2019_BERT}, and is the first large bimodal (NL-PL) pretrained model on CodeSearchNet \citep{Husain2019_CodeSearchNet} dataset. In the pretraining stage, CodeBERT uses two objective functions: Masked Language Modeling (MLM), and Replaced Token Detection (RTD). 

    GraphCodeBERT \citep{Guo2021_GraphCodeBERT} is based on BERT \citep{Devlin2019_BERT}, and is the first pretrained model that leverages the code structure via Data Flow Graph (DFG). In the pretraining stage, GraphCodeBERT uses MLM as a learning objective, and two additional tasks: prediction of code structure edges in the data flow, and variable-alignment over source code and data flow.

    \subsection{Decoder-only models}
    \label{paragraph:decoder_only}

    These methods use decoder(s) only in the Transformer neural network architecture (Figure \ref{fig:transformer_architecture} (b)).

    \paragraph{$\bullet$ PAR: Parametric Loss Function}
    \citep{Hoffmann2022_chinchilla} - This is one of the most recent objectives proposed for decoder-only Chinchilla model. It is a parametric loss function, based on models parameters count and the number of already-seen tokens. The PAR loss is defined in Equation~\ref{eq:PAR}:
    
        \begin{equation}
            \mathcal{L}_{PAR}(\theta, N, D) \triangleq E + \frac{A}{N^\alpha} + \frac{B}{D^\beta}
           \label{eq:PAR}
        \end{equation}
    
    where $E$ is the entropy of the natural language text and refers to ideal generative process, $\frac{A}{N^\alpha}$ refers to the negative gap in performance between a perfectly trained Transformer with N parameters and an ideal generative process, and $\frac{B}{D^\beta}$ refers to incomplete convergence of the Transformer model because of its limited number of optimization steps. The estimation of $\alpha, \beta, A, B, E$ is done using a Huber Loss \citep{hernandez2021scaling} that uses the L-BFGS algorithm \citep{nocedal1980updating}, as in Equation \ref{eq:huber}:
    
        \begin{equation}
            \min_{A, B, E, \alpha, \beta}~ \sum_{runs i}~Huber_\delta \left(log~\mathcal{L}_{PAR}(\theta, N_i, D_i) -  log L_i\right)
           \label{eq:huber}
        \end{equation}

    where $\delta=10^{-3}$ and $L$ is the pretraining loss. Huber loss is robust to outliers and provides an overall good performance.

    \paragraph{$\bullet$ ULM: Unidirectional Language Modeling}  \citep{radford2019_gpt2} - It aims to predict the next token $x_i$ given the previous tokens. The ULM loss function is defined in  Equation~\ref{eq:UML}:

        \begin{equation}
            \mathcal{L}_{ULM}(\theta) = - \sum_{i=0}^{l-1} log~ P_\theta(x_i | x_{<i})
           \label{eq:UML}
        \end{equation}
        where $x_{<i}$ refers to all the tokens that are before the $i^{th}$ one.

    \paragraph{\textbf{\underline{Methods} -}}  
    
     \cite{Lu2021_CodeXGLUE} tested the performance of GPT-2 \citep{radford2019_gpt2} model on code generation task using three configurations: GPT-2, CodeGPT-2 and CodeGPT-adapt. The latter is pretrained on text extracted from 45 million links on the web. Unlike GPT-2, CodeGPT-2 and CodeGPT-adapt that are pretrained on CodeSearchNet dataset, CodeGPT-2 is pretrained from scratch, and CodeGPT-adapt is initialized from GPT-2 checkpoint.

     %GPT-3 (175B parameters)

     \paragraph*{\textbf{Large Language Models (LLMs) - }}
    
    GPT-3 \citep{Brown2020_gpt3} marked the beginning of the era of Large Language Models (LLMs). It showed that, by scaling up language models, the few-shot learning is possible, and it can sometimes even reach competitive results in tasks such as machine translation, code generation, reading comprehension, etc. For all tasks, GPT-3 is applied without any gradient updates or fine-tuning, with tasks and few-shot demonstrations specified purely via text interaction with the model. This model follows GPT-2 architecture, and has 175B of parameters (10 times more than the previous one). 
    
    There is a series of GPT-3.5 models that are based on GPT-3 \citep{Brown2020_gpt3}. We cite them hereafter:
    
    \begin{itemize}
        \item  \textit{Code-davinci-002}: is one of the Codex \citep{Chen2021_codex} models, that are trained on natural language and programming languages including Java.

        \item  \textit{Text-davinci-002}: is an InstructGPT \citep{Ouyang2022_Instruct} model based on code-davinci-002.

        \item  \textit{Text-davinci-003}: is the most powerful GPT-3 model compared with previous GPT-based models. It can perform better in quality language generation, instruction-following, etc.
    \end{itemize}

    An upgrade of GPT-3.5 lead to the birth of ChatGPT~\citep{chatGPT4}, an impressive language model that fueled a debate about the impact of AI on our daily life. Currently, ChatGPT is based on an even more powerful and sophisticated GPT-4 architecture. The latter is a large multimodal model capable of processing both text and image inputs and producing text outputs. It surpasses existing models by a considerable margin in English and also exhibits strong performance in other languages. The loss function computation is one of the main modifications made compared to previous GPT models. The authors were able to predict GPT-4's final loss on their internal codebase by fitting a scaling law with an irreducible loss term. As for Java, ChatGPT is able to generate very complex code from natural language request. 

    Although GPT-4 is an advanced language model that outperforms its predecessors, it still has limitations similar to earlier GPT models. These limitations include generating false information, providing harmful advises, having a restricted context window, and lacking the ability to learn from experience. However, the authors conducted a thorough evaluation of GPT-4's performance on a variety of datasets, such as academic exams. It was found that GPT-4 achieved a human-level performance on most of these tests. Additionally, in their internal factuality evaluations designed to test the model's accuracy, GPT-4 scored 19\% points higher than the latest GPT-3.5 model. Note that ChatGPT (and other large language models) is recent and was not tested yet on CONCODE dataset. For this reason, we did not include its results in our review.

    At the same time than ChatGPT,  \cite{Jack2021_Gopher} built on GPT-2 and proposed Gopher, a model that contains 280 billion parameters. However, Gopher has two modifications compared with GPT-2: the authors used  RMSNorm \citep{zhang2019_RMS} instead of LayerNorm \citep{ba2016_layerNormalization}, and also used the relative positional encoding \citep{dai2019_transformerxl} instead of the absolute positional encodings. The model was evaluated on 152 different tasks, including Java code generation. Even though Gopher outperforms the state of the art in 81\% of the task, when the authors evaluated it in 124 tasks, there are still some challenges related to the distributional bias, and the training definition criteria.
    
    After Gopher, \cite{Hoffmann2022_chinchilla} introduced a new language model called Chinchilla. This model has the same architecture as Gopher but with 70 billion parameters and additional data. Unlike Gopher, Chinchilla uses variations in the number of training steps, model sizes of different training FLOP counts, a different loss function ($PAR$ function from Equation \ref{eq:PAR}), and the AdamW \citep{Loshchilov2019_adamw} optimizer instead of Adam  \citep{Kingma2015_adam}. Chinchilla outperforms various language models such as Gopher and GPT-3 in many tasks. However, the quality of code generation in Chinchilla is related to its size, and scaling training tokens can improve it. Although Chinchilla's authors recommend training a 10B model on 200B tokens, \cite{touvron2023_llama} has shown that a 7B model continues to improve even after 1T tokens.
    
    Inspired by Chinchilla, \cite{touvron2023_llama} proposed the LLaMA family of LLMs, whose parameters range from 7B to 65B. The architecture of LLaMA incorporates the strengths of previous LLMs, such as input normalization inspired by GPT-3 \citep{Brown2020_gpt3}, and SwiGLU activation function from PaLM \citep{Driess2023_PaLM}. Additionally, it employs rotary positional embeddings instead of absolute positional embeddings used by GPTNeo \citep{Black2022_GPTNeo}. LLaMA-13B, which is only one-tenth the size of GPT-3, performs better on most benchmarks and can be run on a single GPU. Furthermore, the 65B-parameter model is competitive with other LLMs like Chinchilla-70B and PaLM-540B.

    \subsection{Encoder-Decoder models}
    \label{paragraph:encoder_decoder}
    
    These methods use both encoder(s) and decoder(s) in the Transformer neural network architecture (Figure \ref{fig:transformer_architecture} (c)).

    \paragraph{$\bullet$ MSP: Masked Span Prediction} \citep{Colin_2020t5}- It consists in randomly masking spans of text with arbitrary lengths. Later, the model predicts the masked spans.MSP loss is defined in Equation~\ref{eq:MSP}.

        \begin{equation}
            \mathcal{L}_{MSP} (\theta) = \sum_{i=0}^{k-1} - log~P_{\theta} (w_{i}^{masked} | w^{\backslash masked}, w_{<i}^{masked})
            \label{eq:MSP}
        \end{equation}
    
        where $x^{masked}$ is the masked input of length $k$,  $x^{\backslash masked}$ is the masked sequence to predict by the decoder, and $x_{<i}^{masked}$ is the span sequence generated so far. 
        
    \paragraph{$\bullet$ IT: Identifier Tagging} \citep{wang-etal-2021-codet5} - It aims to determine if  code tokens are identifiers (e.g., function or variable names). Therefore, authors compute a probability value $p_i$ for each output token.
        
    To see if a node is an identifier or not, the authors convert the code sequence to an AST (Abstract Syntax Tree). The IT loss is computed using a binary cross entropy loss, where the binary label $y_i=1$ means that the token is an identifier, and $y_i=0$ means that it is not (Equation~\ref{eq:IT}).

        \begin{equation}
            \mathcal{L}_{IT} (\theta_e)= \sum_{i=0}^{l-1} - [y_{i}~log~p_{i} + (1-y_{i})~log~(1 - p_{i})]
            \label{eq:IT}
        \end{equation}

        where $\theta_e$ are the encoder's parameters, and $p_{i}$ are the probabilities of the output sequence tokens.
        
    \paragraph{$\bullet$ MIP: Masked Identifier Prediction} \citep{wang-etal-2021-codet5} - It consists in masking all identifiers (function and variable names) in the PL, and replacing all recurrences of a specific identifier with a unique sentinel token. It is inspired by deobfuscation previously used in \cite{Baptiste2021_DOBF}.  The loss function is defined as:

        \begin{equation}
             \mathcal{L}_{MIP} (\theta) = \sum_{i=0}^{n-1} - log~P_{\theta} (w_{i} | w^{masked}, w_{<i})
            \label{eq:MIP}
        \end{equation}

            where  $w^{masked}$ is the masked input.

    \paragraph{$\bullet$ Bimodal dual objective} \citep{wang-etal-2021-codet5} - It aims to leverage bimodal data in order to achieve better NL-PL alignments. During the training process, if NL is the encoder input, then PL is the decoder input and vice versa.

    \paragraph{$\bullet$ MCL: MultiModal Contrastive Learning} \citep{Daya2022_UniXcoder} - This loss helps in learning the semantic embedding of mapped AST sequences. Authors forward the same input in the neural network and use different hidden dropout masks. The resulting hidden representations are used as positives, while the other representations from the same batch are used as negatives.

        \begin{equation}
             \mathcal{L}_{MCL} (\theta) =  - \sum_{i=0}^{B-1} log~\frac{e^{cos(\tilde{h_i}, \tilde{h_i}^+)} / \tau}{\sum_{j=0}^{b-1}e^{cos(\tilde{h_j}, \tilde{h_j^+})} / \tau}
            \label{eq:MCL}
        \end{equation}

    where  $\tau$ is the temperature scalar, $\tilde{h_i}^+$ and $\tilde{h_i}$ are the positive and negative representations, respectively, and $cos (\dot)$ is the cosine similarity.

    \paragraph{$\bullet$ CMG: Cross-Modal Generation} \citep{Daya2022_UniXcoder} - This loss is complementary to the previous one. It pushes the model to provide comments to describe the code's function. This helps the model to better understand the semantics of the code and unify its representation across programming languages. Its is defined in Equation \ref{eq:CMG}.

        \begin{equation}
             \mathcal{L}_{CMG} (\theta) =  - \sum_{i=0}^{l-1} log~P_\theta(d_i | X, d_{t<i})
            \label{eq:CMG}
        \end{equation}

    where X is the flattened AST sequence, and $d = \{d_0, d_1,.., d_{n-1}\}$ is the code comments.

    \paragraph{\textbf{\underline{Methods} -}}   \cite{ahmad-etal-2021-PLBART} introduced PLBART (Program and Language BART), a model based on $BART_{base}$ \citep{lewis-etal-2020-bart}. The latter was originally proposed to tackle text generation and comprehension tasks, such as translation, question answering, and automatic summarization. The only difference between the two model architectures is that PLBART added a normalization layer on top of both the encoder and the decoder. It is the first model to exploit the possibility to generate source code from NL with a pretrained decoder.

    Posterior to PLBART, \cite{phan-etal-2021-cotext} introduced CoTexT (Code and Text Transfer Transformer), which builds on $T5_{base}$ \citep{Colin_2020t5} model. The authors studied the CoTexT performance based on two criteria: (1) combining corpora (\textbf{C}odeSearchNet and \textbf{G}itHub repositories), and (2) pretraining with unimodal and bimodal data. Therefore, the three resulting models are as follows: CoTexT (1-CC), CoTexT (2-CC), and CoTexT (1-CCG). Since all of the models are initialized from $T5_{base}$ weights, and the latter was originally pretrained on the C4 dataset, the comparison is fair. 
    
    Most recently, \cite{lopez2022_JaCoText} introduced JaCoText, a model that is initialized from CotexT weights. It explores additional pretraining using Java code only. In addition, the authors explore the impact of input and output sequence length during fine-tuning.

    \cite{Rizwan2021_RetrievalAugmented} introduced REDCODER (Retrieval augmentED CODe gEneration and summaRization framework). It is composed of a retrieval module called SCODER-R (Summary and CODE Retriever), and a generator module called SCODER-G, which is based on PLBLART \citep{ahmad-etal-2021-PLBART}. SCODER-R is initialized from GraphCodeBERT \citep{Guo2021_GraphCodeBERT} weights, it receives the NL text, and returns the top-\textit{k} source codes. Later, the NL sequence is concatenated with the top-\textit{k} code sequences to have the augmented input sequence. This sequence is fed to PLBART to generate Java code.
    
    CodeT5 \citep{wang-etal-2021-codet5} is based on T5 \citep{Colin_2020t5} neural network architecture. Authors convert PL segments into an Abstract Syntax Tree (AST) and introduce two pretraining tasks: Identifier-aware denoising objective, and bimodal dual generation objective. On the one hand, Identifier-aware denoising objective uses Masked Span Prediction (MSP) objective similar to T5 \citep{Colin_2020t5}, and introduced two supplementary tasks: Identifier Tagging (IT), and Masked Identifier Prediction (MIP). Identifier-aware objective optimizes these three losses (MSP, IT, MIP) equally. On the other hand, bimodal dual objective optimizes the model simultaneously using bimodal data with different order (NL-PL, then PL-NL). Surprisingly, experiments lead by the authors have shown that the order between the code and the natural language test matters in the model's performance.
    
    Later, \cite{sindhu-etal-2022-StructCoder} proposed StructCoder model, which is initialized from CodeT5 \citep{wang-etal-2021-codet5} weights, and pretrained using DAE (Denoising Autoencoding) task. StructCoder also uses an abstract syntax tree and propose two learning objectives to train both the encoder and the decoder: DFG  prediction, and AST paths prediction. Specifically, the decoder is trained to predict the node types on all the root-leaf paths in the target AST, and  to predict the data flow edges.

    Very recently, UniXcoder \citep{Daya2022_UniXcoder} surged to tackle code-related understanding and generation tasks. UniXcoder leverages the AST information through transforming AST to sequence text. This method is functional in three modes: encoder-only, decoder-only, and encoder-decoder. Authors used three learning objective tasks to pretrained the model: MLM \citep{Devlin2019_BERT}, Unidirectional Language Modeling (ULM) \citep{radford2019_gpt2}, and Denoising Objective \citep{Colin_2020t5}. In addition, they introduced two more objectives: Multi-modal Contrastive Learning (MCL), and Cross-Modal Generation (CMG) that target the semantic understanding of code and its unification across different programming languages. 
    
\section{Datasets for Java Code Generation}
    \label{sec:datasets}
    In this section, we present the most used datasets in Java Code Generation. A brief summary of them, as well as their merits and limitations, are presented in Table~\ref{tab:datasets}.

    \paragraph{\textbf{CONCODE}} \citep{iyer-etal-2018-concode} was collected from Java projects on Github (approx. 33,000 repositories). It is composed of tuples of class environment (member variables and member methods), natural language (JavaDoc comments), and source code. It aims to generate Java code from NL documentation and the class environment (used as a context). Java code belongs to several domains. It has been used to fine-tune the models. It contains $100K$, $2K$, and $2K$ lines of NL-PL in the train, validation, and test sets, respectively.

    \paragraph{\textbf{CoNaLa}} \citep{Yin2018_conala} consists of query-code pairs collected from the website Stack Overflow via a mining method. It contains two programming languages: Python and Java. It contains $11125$, $1237$, and $500$ lines of NL-PL in the train, validation, and test sets, respectively.

    \paragraph{\textbf{MTG corpus}} \citep{Ling016_LatentPredictor} aims to generate source code given the card description in Trading Card Games (TCGs). The authors collected the data from two open source implementations of TCGs: Magic the Gathering (MTG), and Hearthtone (HS). While the latter digital implementation is in python. we focus in this work on MTG dataset that contains Java digital implementation. It contains $11969$, $664$, and $664$ lines of NL-PL in the train, validation, and test sets, respectively.
    
   \paragraph{\textbf{CodeSearchNet}} (General Language Understanding Evaluation benchmark for CODE) was introduced by \cite{Lu2021_CodeXGLUE}. It contains six programming languages (Ruby, JavaScript, Go, Python, Java, and PHP) with 6.4M of unimodal (only PL) data, and 2.1M bimodal (PL-NL) datapoints. The data was collected from Github code repositories, and includes 10 tasks over 14 different datasets.  

    \paragraph{\textbf{CodeNet}} was proposed by  \cite{Puri2021_CodeNet}. It contains a set of  programming problems collected from two websites: AIZU \citep{aizu_web} and AtCoder \citep{atcoder_web}. It contains 13,916,868 submissions, divided into 4053 problems. Submissions are in 55 different programming languages, but 95\% are coded in C++, Python, Java, C, Ruby, and C\#.  Only the other 5\% are in Java.  
    
    \paragraph{\textbf{Java dataset}} \citep{ahmad-etal-2021-PLBART} was collected from Github repositories associated with Java language available on Google BigQuery. Authors follow \cite{Lachaux2020_Unsupervised} to preprocess the data.  

    \paragraph{\textbf{AlphaCode Pre-training}}   was introduced by \cite{li2022_AlphaCode} to pretrained their model. The authors scrawled the data from public Github repositories. The pretrained dataset contains 715.1 GB of code, including  C++, C\#, Go, Java, JavaScript, Lua, PHP, Python, Ruby, Rust, Scala, and TypeScript. From these programming languages, 113.8 GB consist of Java data. 

    \paragraph{\textbf{CodeContests}}  was introduced by \cite{li2022_AlphaCode} to fine-tuned their model. The authors combines scraped data from Codeforces \citep{Codeforces2020} with existing data from Description2Code \citep{Caballero_Description2Code_Dataset_2016}, and CodeNet \citep{Puri2021_ProjectCodeNet}. It contains C++, Python, and Java.
    
    \paragraph{\textbf{BIGQUERY CODEGEN}} \citep{Pang2022_CodeGen}  contains six programming languages: C, C++, Go, Java, JavaScript, and Python. In total, it contains, 342.1 GB of data, including around 120.2 GB of Java data. It was used to pretrain the CODEGEN models.

    \paragraph{\textbf{PolyCoder Dataset}} \citep{Xu2022_PolyCoder}  was collected from Github repositories that have at least 50 stars. It contains 12 PLs (C, C\#, C++, Go, Java, JavaScript, PHP, Python, Ruby, Rust, Scala, TypeScript), where each  PL has a maximum of 25K repositories. The raw data size is around 631 GB (including 60 GB for Java). To preprocess the data, the authors follow Codex  process \citep{Chen2021_EvaluatingLLCodex}. After filtering the data, the final data size is 243.6 GB, with 41 GB for Java language.

    \paragraph{\textbf{The Stack Dataset}}  was proposed by \cite{Kocetkov2022_Stack3B}. It is the largest code dataset so far. It contains 6 TB of data from 358 programming languages. The code was extracted from Github repositories with permissive licenses, such as MIT, BSD-3-Clause, ISC, ECL and Apache 2.0.

\begin{table*}\centering
\begin{center}
\resizebox{\textwidth}{!}{
    \begin{tabular}{|c|c|c|c|c|c|c|}
    \toprule
    Dataset & \makecell{Programming\\ language (PL)} & Size & Source & Proposed for & merits & limitations\\
    \midrule
    CONCODE & Java & \makecell{$100K$, $2K$, $2K$ lines\\ in train/val/test} & Github & code generation & large scale & \makecell{query nl incorporates both member \\ variables  and member methods.  However, \\ in practice, is not an optimal approach.} \\    \hline
    
    CoNaLa & Python, Java & \makecell{$11125$, $1237$, and $500$ lines\\ in train/val/test} & Stack Overflow & classification & manual annotation & \makecell{- If there are several valid source code \\  for a question, the annotators \\ might not identify all of them. \\ - The annotators can fail when the \\ code source is  complex. } \\    \hline
    
    MTG corpus & Python, Java & \makecell{$11969$, $664$, and $664$ lines\\ in train/val/test}  & \makecell{trading card games \\Magic the Gathering\\ and Hearthtone} & \makecell{code generation}  & \makecell{new area of focus on \\ code generation \\ for card games.} & \makecell{- The programming of card  games is \\ a niche industry that caters to a \\ specific demographic within society.} \\    \hline
    
    CodeSearchNet & \makecell{Ruby, JavaScript, Go,\\Python, Java, PHP} & \makecell{6.4M of PL data,\\2.1M of PL-NL data} & Github & \makecell{code search} & \makecell{- large scale\\- expert annotators} & \makecell{The authors create proxy dataset\\ of lower quality} \\  \hline
    
    \makecell{BIGQUERY\\CODEGEN} & \makecell{C, C++, Go,\\Java, JavaScript, Python} & \makecell{342.1 GB of total code,\\120.3 GB of Java code} & - & \makecell{Multi-Turn \\program synthesis} & \makecell{- large scale\\- multi-lingual dataset} & -  \makecell{The code in the dataset \\ exhibits certain vulnerabilities \\ and safety issues.} \\    \hline
    % MULTI-TURN PROGRAM SYNTHESIS
    
    CodeNet & \makecell{55 different \\ programming languages} & \makecell{14 million of code \\ samples and approx. 500 \\ million lines of code} & AIZU and AtCoder & \makecell{- code similarity \\ and classification \\ code translation\\ - code performance } & \makecell{- large scale\\- high-quality \\annotations} & \makecell{ - The code samples may lack \\ extensive comments and could \\ be in multiple languages. \\ - The code samples offer \\ solutions for programming problems \\ that are aimed at beginners  \\in college and high school. }\\  \hline
    
    \makecell{AlphaCode\\Pre-training \\ dataset} & \makecell{C++, C\#, Go,\\Java, JavaScript, Lua,\\PHP, Python, Ruby,\\Rust, Scala, TypeScript} & \makecell{715.1 GB of total code,\\113.8 GB of Java code} & Github  & code generation & \makecell{- large scale\\- multiple PLs} & \makecell{The dataset may contain \\ unsafe code} \\    \hline

    \makecell{CodeContests} & \makecell{C++, Python, Java} & \makecell{13328, 117, 165 in lines \\ train/valid/test} & \makecell{Codeforces \\ Description2Code \\ CodeNet}  & code generation & \makecell{- includes problems, \\ solutions and test cases \\ - contains correct and \\ incorrect human \\ submissions} & \makecell{- the raw datasets of \\  CodeContests contain a  \\ high  rate of false \\ positive samples.} \\    \hline

    \makecell{PolyCoder\\Dataset} & \makecell{C, C\#, C++,\\Go, Java, JavaScript,\\PHP, Python, Ruby,\\Rust , Scala, TypeScript} & \makecell{631 GB of total code,\\60 GB of Java code} & Github & \makecell{code completion \\code synthesis} & \makecell{- large scale\\- multiple PLs} & \makecell{The code source may \\ contain vulnerabilities \\ and safety issues.} \\    \hline
    
    \makecell{The Stack\\Dataset} & \makecell{358  languages\\including Java} & \makecell{6 TB  of total code}  & Github & \makecell{-  code completion \\ - documentation generation \\ - auto-completion of \\ code snippets} & \makecell{- large scale\\- multiple PLs} & \makecell{it contains unsafe code\\ (malicious files)} \\  \hline
    
    Java dataset & Java & 352 GB & \makecell{Github and \\ StackOverflow} & \makecell{- code generation} & \makecell{- specialized in Java} & \makecell{- the code may contain \\vulnerabilities} \\  
    
    \bottomrule
    \end{tabular}
}
\end{center}
    \caption{Summary of datasets used in literature}
    \label{tab:datasets}
\end{table*}

\section{Evaluation metrics}
\label{sec:performanceMetrics}
    The following metrics are used to evaluate Java Code Generation methods. A brief summary of these metrics is presented in Table~\ref{tab:metrics}.
    
    \paragraph{\textbf{Exact Match (EM)}} measures the accuracy of the model. It determines if the prediction is the same as the ground truth (also known as reference or gold standard). Even though this metric is easy and fast to compute, and provides straightforward understanding of model's accuracy, it is too rigid that it does not take into account the semantic similarity between reference and prediction codes. It penalizes the model even for white spaces and the variable names. Naturally, this is the most difficult metric to satisfy.
    
    \paragraph{\textbf{BLEU}} \citep{papineni2002_bleu} was originally introduced to evaluate machine translation task. It measures the percentage of overlapped n-grams between the ground truth and the prediction generated by the model. It considers some criteria to penalize the score, such as the sentence length. Overall, this metric is less rigid than EM, and accepts small variations between two predicted codes. However, since the Ngrams order is not considered, it also ignores the semantic of the generated code. Also, it can overestimate the model's score.

    \paragraph{\textbf{CodeBLEU}} \citep{Ren2020_CodeBLEU} uses the n-gram match score of the BLEU metric between the ground truth and the generated sequence. Furthermore, it takes into account the code syntax, and code semantics by matching the AST and DFG, respectively. This metric is the most accurate one so far and is better adapted for code generation. Unlike BLEU metric, it can be too strict and underestimate the model's score.

\begin{table*}\centering
\begin{center}
\resizebox{\textwidth}{!}{
    \begin{tabular}{|l|c|c|c|c|}
    \toprule
    Metric & \makecell{Proposed for} & Merits & Limitations\\
    \midrule
    Exact Match (EM)  & \makecell{- code completion\\- code summarization}  & \makecell{- easy to compute\\- fastest in inference\\- easy to interpret}  & \makecell{- rigid metric\\- semantic not considered\\- penalizes even white spaces}  \\
    \hline
    BLEU \citep{papineni2002_bleu} & machine translation  &  \makecell{- less rigid that EM\\- easy to compute\\- accepts multiple correct answers} &  \makecell{- Ngram order not considered\\- semantic not considered\\- can be too optimistic}  \\ \hline
    CodeBLEU \citep{Ren2020_CodeBLEU}  & code generation & \makecell{- more accurate\\- better adapted for code generation\\- better handling of semantic}  & \makecell{- slower in inference\\- more difficult to compute\\- can be too strict}  \\ 
    \bottomrule
    \end{tabular}
}
\end{center}
    \caption{Summary of metrics used in literature}
    \label{tab:metrics}
\end{table*}

\begin{table*}\centering
\begin{center}
\resizebox{0.85\textwidth}{!}{

    \begin{tabular}{|c|c|c|c|c|c|c|c|}
    \toprule
    Model & \makecell{Input / Output\\ Length} & LR & Optimizer & Dropout & \makecell{Batch \\ Size (B)} & \makecell{Training \\ steps} & \makecell{Vocab \\ Size}\\
    \midrule
    GPT-2  pretraining  & 1024  & 5e-5 &  Adam  & 0.1 & 32 & - &  50,000 \\
    \hline
    PLBART pretraining & 512 /512 & 1e-6 & Adam & 0.1, 0.5  & 2048 & 100K & 50,004\\
    PLBART fine-tuning  & 512 /512 & 3e-5 & Adam & 0.1       & 32   & 100K & 50,004 \\ 
    \hline 
    CoTexT pretraining      &1024 /1024  & 1e-3 & Adam & - & 128 & 200K & 32,000  \\
    CoTexT-base fine-tuning  & 256 / 256  & 1e-3 & Adam & - & 128 & 45K  & 32,000 \\ 
    \hline
    JaCoText-base fine-tuning  & 379 / 379  & 1e-3 & Adam & - & 128 & 60K  & 32,000 \\ 
    \hline
    CodeBERT fine-tuning      & 512 &  5e-5  & Adam & 0.1 & 32 &  & 50,265  \\ 
    GraphCodeBERT fine-tuning & 256 &  1e-4  & Adam &  -  & 64 &  &  -  \\
    \hline

    CodeT5-base pretraining & 512 / 256 & 2e-4 & Adam & - & 1024 & 150 epochs & 32,100  \\
    CodeT5-base fine-tuning  & 320 / 150 & 5e-5 & Adam & - &  32  &  30 epochs & 32,100  \\ 
    \hline
    StructCoder pretraining & 400 / 400 & 5e-5 & AdamW  &  -  & 32  &  12K  & 32,100 \\
    StructCoder fine-tuning  & 325 / 155 & 5e-5 & AdamW  &  -  & 32  &  100K & 32,100 \\
    
    \bottomrule
    \end{tabular}
}
\end{center}
    \caption{Hyper-parameters used by state-of-the-art methods during pretraining and fine tuning}
    \label{tab:hyperparams}
\end{table*}

\section{Results and Discussion}
\label{sec:results}

    \paragraph*{\textbf{Main Results - }}

    Table \ref{tab:hyperparams} displays the hyper-parameters utilized by state-of-the-art techniques during the pretraining and fine-tuning stages to evaluate the CONCODE dataset. Meanwhile, Table~\ref{tab:java_results_concode} presents the outcomes of diverse approaches on the same dataset. Since CONCODE was released in 2018, it has been the main benchmark to evaluate Java code generation task. The first two lines of the table correspond to results of RNN-based methods. The latter obtain the lowest scores. This finding is not surprising because recurrent neural networks are not as large as Transformers, and have limited generalization capability. In addition, they suffer often from vanishing and exploding gradient problems \citep{hochreiter1998_vanishing,squartini2003_Vanishing, fadziso2020_overcoming}.

    As for the Transfomer-based methods, most of encoder-decoder methods are based on T5 (namely CodeT5, CoTexT, JaCoText, and StructCoder) or BART  (namely PLBART) models. In the case of T5-base, the scores are obtained after fine-tuning directly from the pretrained model. However, PLBART, CoTexT and JaCoText were pretrained using source code datasets, and are based on previously trained models.

    Encoder-only methods are generally based on BERT. Authors of CodeBERT and GraphCodeBERT did not originally used them for Java Code Generation on CONCODE dataset. Instead, they used them in other tasks such as code search and code documentation generation. Results presented here were reported by \cite{Rizwan2021_RetrievalAugmented} who used the models for code generation. Note that CodeBERT \citep{feng2020_codebert} is the first model to be pretrained on various programming languages.

    Decoder-only  models presented by \cite{Lu2021_CodeXGLUE}  are versions based on GPT-2 \citep{radford2019_gpt2}. The first model is directly fine-tuned from GPT-2 weights, while the second and third models were trained on the CodeSearchNet dataset. The main difference is that CodeGPT-adapted initializes the training process from the GPT-2 checkpoint, and CodeGPT-2 performs a training from scratch. 

    Results show that performance of the Transformer-based models varies significantly. The highest overall results are obtained with the very recent state-of-the-art methods that are based on T5 \citep{Colin_2020t5} and GPT-2 \citep{radford2019_gpt2} models. The best model in terms of BLEU and CodeBLEU metrics is CodeT5-large. This is intuitive insofar as it has the largest number of parameters (770 M). However, this boost in performance comes at an expense of longer training time due to the large complexity of the model. The best model in terms of Exact Match metric is REDCODER \citep{Rizwan2021_RetrievalAugmented}, indicating that it generates code that is more similar syntactically to the ground-truth code. It is noteworthy to mention that, on the one hand, results are very comparable between CodeT5-large and REDCODER, while on the other hand, there is a huge difference between their number of parameters. In fact, CodeT5-large is x5.5 larger than REDCODER. Depending on the application domain and on the available computational resources, a compromise should be done in order to prioritize the gain in memory or the gain in performance.

    CodeBERT \citep{feng2020_codebert} and GraphCodeBERT \citep{Guo2021_GraphCodeBERT} have also achieved relatively high scores, but not as high as the CodeT5-large and REDCODER. This is likely due to the fact that they have fewer number of parameters (125M, 110M versus 770M and 140M, respectively). Interestingly, REDCODER outperforms CodeGPT-2 even thought it has less than half of its parameters only (140 M vs 345 M). This proves the effectiveness of retrieval and generation models SCODER-R and SCODER-G, the main building blocks of REDCODER.

    StructCoder \citep{sindhu-etal-2022-StructCoder} and JaCoText \citep{lopez2022_JaCoText} have very comparable EM (22.35 vs 22.15) and BLEU (40.91 vs 39.07) results, with StructCoder being slightly better than JaCoText. The latter improves the results of CoText by an average of 2.5 points. JaCotext increments CotexT by adding a pretraining phase in which large input sequence were used. Finally, UniXcoder, CodeT5-small, and $T5_{base}$ achieve the lowest results in Transformer-based methods. Overall Code-T5-small is the model that best balances the compromise between performance and number of parameters, and is preferable is stricter application domains.

    Last but not least, results show that using pretraining leads to a better performance than training from scratch. This can be seen when comparing the performances of CodeGPT-adapted and CodeGPT-2. As mentioned previously, the former is initialized from GPT-1 checkpoints, and the latter is trained from scratch.
    This finding was also confirmed by \cite{feng2020_codebert} and \cite{Baptiste2021_DOBF}. Therefore, we recommend using pretrained models whose representation is already stable and can facilitate the training of down-streaming tasks.
    
    Finally, since the highest achieved score on CONCODE dataset is 45.08, there is still a large gap between the best Java code generation methods and the ground truth. This means that more effort is needed to tackle this task.

\begin{table*}[!htb]
\centering
\resizebox{0.85\textwidth}{!}{
\begin{tabular}{|c|c|lccc|c|}
\hline
\multicolumn{3}{|c}{} & \multicolumn{3}{c|}{Score  ($\uparrow$}) & \# params ($\downarrow$)\\ \hline
\multicolumn{2}{|c|}{\textbf{Category}} & \textbf{Model} & \textbf{EM} & \textbf{BLEU} & \textbf{CodeBLEU} &  \\ \hline
& \multirow{2}{*}{RNN-based} & Seq2Seq \citep{yin2017_syntactic} & 6.65 & 21.29  & - & - \\ \cline{3-7}
&  &\cite{iyer-etal-2018-concode} & 8.60 & 22.11 & - & - \\
\hline

\multirow{2}{*}{\rotatebox[origin=c]{90}{Transformer-based Methods~~~~~~~~~~~~~~~}} & \multirow{2}{*}{Encoder-only} &CodeBERT \citep{feng2020_codebert} & 18.00  &  28.70 & 31.40 & 125M \\\cline{3-7}
&  & GraphCodeBERT \citep{Guo2021_GraphCodeBERT} & 18.70 & 33.40 & 35.90 & 110M \\

\cline{2-7}

& \multirow{3}{*}{Decoder-only} &GPT-2 \citep{radford2019_gpt2}  & 17.35 & 25.37 & 29.69 & 345M  \\\cline{3-7} 
&  & CodeGPT-2 \citep{Lu2021_CodeXGLUE} & 18.35 & 28.69 & 32.71 & 124M \\\cline{3-7}
&  &\small CodeGPT-adapted \citep{Lu2021_CodeXGLUE} & 20.10 & 32.79 & 35.98 & 124M  \\
\cline{2-7}
& \multirow{13}{*}{Encoder-decoder} &$T5_{base}$ \citep{Colin_2020t5} & 18.65 & 32.74 & 35.95 & 220 M  \\ \cline{3-7}
&  &CodeT5-small \citep{wang-etal-2021-codet5} & 21.55 & 38.13 & 41.39 &   \\ 
&  &+dual-gen   & 19.95  & 39.02 & 42.21 & 60 M \\ 
&  &+multi-task & 20.15 & 35.89 & 38.83 &   \\
\cline{3-7}
&  &CodeT5-base \citep{wang-etal-2021-codet5} & 22.30 & 40.73 & 43.20 &  \\
 & &+dual-gen & 22.70 & 41.48 & 44.10 & 220 M \\ 
&  &+multi-task & 21.15 & 37.54 & 40.01 & \\\cline{3-7} 
&  &CodeT5-large \citep{Le2022_codeRL} & 22.65 & \textbf{42.66} & \textbf{45.08} & 770 M  \\\cline{3-7}
&  &PLBART \citep{ahmad-etal-2021-PLBART}  & 18.75 & 36.69 & 38.52 & 140 M \\ \cline{3-7}
&  &CoTexT \citep{phan-etal-2021-cotext} & 20.10 & 37.40 & 40.14 & 220 M \\\cline{3-7}
&  &JaCoText \citep{lopez2022_JaCoText} & 22.15 & 39.07 & 41.53 & 220 M  \\\cline{3-7}
&  &REDCODER \citep{Rizwan2021_RetrievalAugmented} & \textbf{23.40} & 41.60 & 43.40 & 140 M \\\cline{3-7}
&  &REDCODER-EXT \citep{Rizwan2021_RetrievalAugmented}& 23.30  & 42.50 & 43.40 & 140 M \\\cline{3-7}
&  &StructCoder \citep{sindhu-etal-2022-StructCoder}  & 22.35 & 40.91 & 44.76 & 220 M \\ \cline{3-7}
&  &UniXcoder \citep{Daya2022_UniXcoder} & 22.60  & 38.23 & - & 125M \\ 

\hline
\end{tabular}
}
\caption{Results of Java code generation task on the CONCODE dataset. We present EM, BLEU, and CodeBLEU scores, as well as the number of parameters for each method. Best results are in bold.}
\label{tab:java_results_concode}
\end{table*}

\paragraph*{\textbf{General Discussion - }}

    Over the years, generating code through natural language has been a challenging task. The main goal of this task is to increase programmers' productivity by automating tedious and repetitive coding tasks. In this paper, we present advancements in code generation within the Java programming language. However, in order to fully comprehend the subject matter, it is crucial to have a grasp of the historical development of code generation in other fields.
    
    Early attempts to generate code automatically involved translating natural language requests into regular expressions \citep{Angluin1987_LearningRegular,Ranta1998_Multilingual}, logical forms based on probabilistic categorial grammars \citep{Zettlemoyer2055_LogicalForm}, agent-specific language \citep{Kate2005_LearningTransform}. These approaches laid the groundwork for more sophisticated techniques, such as semantic parsing \citep{wong2006_LearningSemantic, lu2008_generativeModel} and SQL query generation \citep{miller1996_FullyStatistical, Ramaswamy2000_HierarchicalFT}, which were discussed in Section \ref{sec:background}.
    
    The previously mentioned methods utilize basic techniques such as rule-based systems and simple statistical methods to overcome the different tasks. However, as time has progressed and technology has advanced, the demand for more sophisticated techniques has increased to tackle the most complex code generation obstacles. One of the techniques that has made significant strides in this field is the utilization of RNNs, such as LSTMs \citep{hochreiter1997_lstm}, and later GRUs \citep{cho2014_gru}. These advanced techniques have greatly enhanced the capability to address the challenges of code generation. For instance, \cite{iyer-etal-2018-concode} introduced CONCODE, a dataset with the purpose of generating Java code through natural language. The authors used a sequence-to-sequence model based on RNNs, which was similar to other models used in the field \citep{Neelakantan2016_NeuralProgrammer,Ling016_LatentPredictor}. At the time of the development of this model, the integration of the attention mechanism \citep{Gu2016_CopyingMechanism} was a significant breakthrough in code and text generation. 

    Although models based on Recurrent Neural Networks (RNNs) were initially promising, they were hindered by several drawbacks. One of these was the vanishing gradient problem \citep{hochreiter1998_vanishing,squartini2003_Vanishing, fadziso2020_overcoming}, and also they were limited in their ability to effectively process large amounts of input data. As a result, RNN-based models often truncated natural language requests and code sequences, leading to significant information loss and incomplete generated code. Furthermore, due to Java's strict programming rules, the code generated by these models was often riddled with syntax errors and could not be executed.

    After the big success of Transformers \cite{vaswani2017_attention} in NLP tasks, many Java Code Generation researchers shifted their efforts to this neural network architecture, and proposed the above mentioned methods. The most succesful models are based on T5 and GPT-2 models. CodeT5 models use a learning objective task that was specifically designed to take advantage of the programming language structure. In \cite{wang-etal-2021-codet5}, CodeT5-small and CodeT5-base achieved remarkable improvements in the performance of code generation task using bimodal dual learning objective. Authors leverage the token type of identifiers from the AST such as function names and variables to enrich the semantic features. \cite{wang-etal-2021-codet5} studied three objectives: Masked Span Prediction (MSP),  Identifier Tagging (IT), and  Masked Identifier Prediction (MIP) . It turns out that Masked Span Prediction is the most crucial objective for learning the syntactic information in the generation tasks. However, the simultaneous optimization of the three of them achieve the best performance. Alternatively, \cite{Le2022_codeRL} show that CodeT5-large obtains higher scores than CodeT5-small and CodeT5-base. This finding is intuitive since the model architecture is more complex.

    After the success of CodeT5, StructCoder \citep{sindhu-etal-2022-StructCoder} surged. It combines T5-base architecture with ASTs. Moreover, authors use DFGs to make aware both the encoder and decoder of the syntax and the data flow. This system achieved better results than CodeT5 using T5-base. Therefore, AST and DFG objectives are beneficial to learn correctly the semantic and syntax of Java programming language. Another interesting models is REDCODER, the latter is the first model to use two encoders. The first encoder is used for the retrieval module to search for relevant code in the dataset, the second one is part of the  generator module (encoder-decoder model) to generate the Java source code. REDCODER  achieves the best Exact Match (EM) score.

    GPT-3 model in an improvement of GPT-2, introduced by \cite{Brown2020_gpt3}. It is a breakthrough in Large Language Models (LLMs), as it demonstrates the feasibility of scaling the model and achieving competitive scores in various tasks through few-shot learning. This model was the starting point to many variants among which OpenAI's ChatGPT chatbot \citep{chatGPT4}. ChatGPT \citep{chatGPT4} is based on the GPT-3.5 model, which has shown remarkable performance in various tasks, such as machine translation in multiple languages, automatic summarization, parts of speech tagging, and even code generation through natural language requests. In particular, ChatGPT has demonstrated impressive proficiency in writing complex Java source code, even for applications.

    Despite the impressive results showed by ChatGPT, the latter has many limitations, out of which, we mention the following. (1) ChatGPT is not able to fully understand natural language requests that are not well-formulated (such as missing commas), and (2) ChatGPT is sensitive to the prompts used by users to interact with it. The new version of ChatGPT (called ChatGPT4) was proposed precisely to augment the understanding capacity of it. ChatGPT4 has made significant advancements in Java Code generation from complex queries, as well as in working on multimodal inputs such as text, images, and video. 

    ChatGPT4 is not the only attempt to improve ChatGPT, other decoder-only competitors are trying hard to outperform it. Two examples of such systems are Google's BARD \citep{Google_Bard} that is based on LaMDA model \citep{Thoppilan2022_LaMDA},  and LLaMA \citep{touvron2023_llama} that is based on Chinchilla model. It is noteworthy to mention that all very recent LLMs have not been yet fully benchmarked on any of state-of-the-art datasets for Java code generation, at the time of writing this article.

\section{Conclusions and Perspectives}
\label{sec:Conclusions}

In this paper, we presented a comparison review of state-of-the-art methods in Java Code Generation using CONCODE dataset. Many methods surged during the last years and we categorized them based on their architecture: (1) RNN-based methods, and (2) Transformer-based methods. The latter are divided into three subcategories: encoder-based methods, decoder-based methods, and encoder-decoder-based methods. We propose the following future directions:
    
    \begin{itemize}
        \item It would be interesting to use beam search algorithm to detect syntax errors, similarly to what is done in the semantic parser PICARD in SQL code generation. \citep{Scholak2021_PICARD}

        \item One big drawback of current code generation metrics is their high dependency on syntax similarity between reference and predicted codes. It is really important and right time to start focusing on handling the semantic part of it. Even though that two codes are syntactically different, they still can perform the same task.
        
        \item One straightforward line of work is to exploit more Java datasets such as AlphaCode \citep{li2022_AlphaCode}, PolyCoder \citep{Xu2022_PolyCoder}, and AlphaCode \citep{li2022_AlphaCode} in the generation process, especially at the fine tuning step. This is because current methods only use Java data to pretrain the model, while they fine-tune models on Python benchmarks. 

        \item Despite the fact that large language models such as ChatGPT, Bard and LLaMA provide extraordinary performances in code generation, their number of parameters is huge. Future research would focus on the compression of such language models, or the proposal of smaller-scale efficient language models. This will be very useful for researcher and companies that do not have enough computational resources or memory \citep{tao2022_compression}.
        
        \item 
        Alternatively, another perspective would be to work on training time optimization, especially for large models such as ChatGPT. For instance, continual learning algorithms can be used in order to integrate new data on-the-fly, without retraining the model from scratch at each update \citep{gao2023_continual}.

        \item Current code generation models lack of reasoning abilities. Another perspective would be to develop more the common sense of language models to improve their generation semantic, thus, their performance \citep{huang2022towards}.

    \end{itemize}

\bibliographystyle{cas-model2-names}
\bibliography{cas-refs}
\end{document}